\documentclass[runningheads]{llncs}
\usepackage[T1]{fontenc}
\usepackage{graphicx}
\usepackage{ifthen}
\usepackage{color}
\usepackage{amsmath}
\usepackage{amssymb,amsfonts}
\usepackage{psfrag}
\usepackage{xcolor} 
\usepackage[ruled,vlined]{algorithm2e}
\usepackage{multirow}

\usepackage{url}

\usepackage{appendix}

\begin{document}
\title{One-Class Domain Adaptation via Meta-Learning}

%
%
\author{Stephanie Holly\inst{1} \and
Thomas Bierweiler\inst{3} \and
Stefan von Dosky\inst{3} \and Ahmed Frikha\inst{2} \and Clemens Heitzinger\inst{1} \and Jana Eder\inst{3}}
\authorrunning{S. Holly et al.}
%
\institute{Vienna University of Technology
 \and Ludwig Maximilian University of Munich \and
Siemens AG}

\maketitle              
\begin{abstract}
The deployment of IoT (Internet of Things) sensor-based machine learning models in industrial systems for anomaly classification tasks poses significant challenges due to distribution shifts, as the training data acquired in controlled laboratory settings may significantly differ from real-time data in production environments. Furthermore, many real-world applications cannot provide a substantial number of labeled examples for each anomalous class in every new environment. It is therefore crucial to develop adaptable machine learning models that can be effectively transferred from one environment to another, enabling rapid adaptation using normal operational data. We extended this problem setting to an arbitrary classification task and formulated the one-class domain adaptation (OC-DA) problem setting. We took a meta-learning approach to tackle the challenge of OC-DA, and proposed a task sampling strategy to adapt any bi-level meta-learning algorithm to OC-DA. We modified the well-established model-agnostic meta-learning (MAML) algorithm and introduced the OC-DA MAML algorithm. We provided a theoretical analysis showing that OC-DA MAML optimizes for meta-parameters that enable rapid one-class adaptation across domains. The OC-DA MAML algorithm is evaluated on the Rainbow-MNIST meta-learning benchmark and on a real-world dataset of vibration-based sensor readings. The results show that OC-DA MAML significantly improves the performance on the target domains and outperforms MAML using the standard task sampling strategy.

\end{abstract}
\section{Introduction}
In recent years, the integration of Internet of Things (IoT) sensor platforms into industrial plants has opened up new avenues for the application of machine learning models in various industrial systems \cite{Kemnitz2022,Heistracher2021,Kumar2021,Holly2021}, offering significant opportunities for enhancing efficiency in industrial processes. However, deploying machine learning models in such systems poses significant challenges, particularly in ensuring consistent performance across diverse environments within an industrial plant. In many industrial machine learning applications, models are developed in controlled laboratory settings before being deployed into critical production environments \cite{Eder2023}. However, real-world data is inherently complex and diverse, and thus, the training data may significantly differ from the real-time data in production. Distribution shifts -- where the training distribution differs from the test distribution -- pose significant challenges in the application of machine learning, as they can lead to a substantial decline in model performance \cite{Koh2021}. Therefore, it is crucial to develop robust machine learning models that can be effectively transferred from laboratory settings to real-world deployments, as well as from one environment to another within an industrial system.

The generalization ability of machine learning models has been significantly driven by the wealth and diversity of available data. This is evident in computer vision, where datasets like ImageNet \cite{Russakovsky2014} have improved the performance of image classification models. Although these pre-trained models can be effectively applied to related computer vision tasks, such extensive datasets are often unavailable in many specialized domains \cite{Hospedales2020}. Real-world applications often face significant limitations in data availability. This scarcity is often due to intrinsic factors (e.g. rare medical conditions, special industrial failure types) or the time-consuming and costly nature of data acquisition processes. Furthermore, the adaptation data in new domains is constrained not only by limited volume but also by the scarcity of certain classes. For illustration, machine learning models in industry are typically pre-trained in laboratory settings before being transferred to real-world deployments. While normal operational data is often readily available for model adaptation, acquiring anomalous data within large and complex industrial systems is both expensive and time-consuming \cite{Frikha2021}. This raises the question of whether and how the information present in normal operational data can be leveraged for adaptation to a new domain. 



We extended this problem setting to an arbitrary classification task and formulated the \textit{One-Class Domain Adaptation} (OC-DA) problem setting. To the best of our knowledge, our work is the first to address one-class adaptation across domains. The goal of OC-DA is to learn a model that is able to quickly adapt to a new domain using only a few examples of \textit{one} class. We take a meta-learning approach to tackle the challenge of OC-DA and propose a task sampling strategy to adapt any bi-level meta-learning algorithm to the OC-DA setting. In summary, our contributions are as follows: (1) We formulate a new problem setting in the context of domain adaptation (OC-DA) motivated by real-world challenges and requirements. (2) We propose a sampling strategy to adapt any bi-level meta-learning algorithm to OC-DA. We focus on modifying the MAML algorithm and introduce the OC-DA MAML algorithm. (3) We provide a theoretical analysis showing that the OC-DA MAML algorithm optimizes for meta-parameters that enable rapid one-class adaptation across domains. (4) We empirically evaluate the OC-MAML algorithm on a meta-learning benchmark, the Rainbow-MNIST dataset \cite{Finn2019}, and on a dataset of sensor readings recorded by centrifugal pumps within diverse environments, demonstrating its robustness for real-world applications.


\section{Methodology}
\subsection{Related Work}
There are two common approaches to tackle the challenge of distribution shifts: domain generalization (DG) and domain adaptation (DA). DG methods train a model on the source domains, aiming to ensure the model performs well in a new target domain without further adaptation \cite{Li2017,Frikha2021b}. Unlike standard algorithms that focus on minimizing the empirical risk (ERM), DG methods frequently incorporate a penalty term to promote invariance across domains \cite{Koh2021}. For instance, invariant risk minimization (IRM) aims to learn an invariant model across multiple domains with the ability to generalize to new domains by learning a data representation, typically through a neural network, such that the optimal classifier built upon this data representation is consistent across all source domains \cite{Arjovsky2019}. In contrast to DG, DA methods utilize unlabeled or sparsely labeled data in the target domain to adapt the model. In transfer learning, a model is initially pre-trained on source domains using standard learning techniques and subsequently fine-tuned on a smaller dataset from the target domain. However, the effectiveness of fine-tuning diminishes significantly when the target domain dataset is very small \cite{Howard2018}. 

Meta-learning, on the other hand, is specifically tailored to few-shot learning settings and can effectively address both DA and DG tasks \cite{Hospedales2020,Li2017,Li2020,Frikha2020}. Meta-learning is commonly understood as 'learning to learn', referring to the process of improving a learning algorithm \cite{Hospedales2020}. There are three common approaches to meta-learning: metric-based, model-based and optimization-based meta-learning \cite{Hospedales2020}. These approaches differ in how they leverage the support set $S$ to model the predicted probability $p_{\theta}(y|x, S)$. Metric-based meta-learning focuses on learning a similarity measure $k_{\theta}$ to compare new examples with those in the support set \cite{Koch2015,Snell2017,Vinyals2016}. Model-based meta-learning directly trains models to represent $p_{\theta}(y|x, S)$. Optimization-based meta-learning aims to optimize the learning process itself by learning an adaptation procedure $\mathcal{A}lg(\theta, S)$ \cite{Finn2017,Rajeswaran2019,Nichol2018,Ravi2017}. The adaptation procedure computes effective task-specific parameters using the examples from the support set. 

Our approach is related to previous work by Li et al. \cite{Li2017} that proposes a meta-learning method for domain generalization (MLDG). Rather than explicitly training for rapid adaptation, MLDG explicitly trains a model for good generalization ability to new domains. MLDG simulates domain shifts by splitting the source domains into training and virtual testing domains. The key idea is that gradient descent steps on training domains should also improve performance on virtual testing domains. The most related work to ours is the task sampling strategy in few-shot one-class classification (FSOC) by Frikha et al. \cite{Frikha2021}, studying the intersection of few-shot learning and one-class classification. One-class classification refers to learning a binary classifier that can differentiate between in-class and out-of-class examples using only in-class data \cite{Frikha2021}. The goal of FSOC is to learn a binary classifier such that fine-tuning on a few in-class examples achieves the same performance as doing so with a few in-class and out-of-class examples. The class-imbalance rate indicates the ratio of in-class and out-of-class examples. FSOC modifies the task sampling strategy in bi-level meta-learning algorithms such that the class-imbalance rate in the support set of meta-training tasks matches the one in the support set of meta-testing tasks, while the query set of meta-training and testing tasks is class-balanced. 


\subsection{Meta-Learning}
In supervised meta-learning, a dataset contains pairs of feature vectors $x$ and labels $y$. The training dataset $D^{\text{train}}$ is divided into a collection of meta-training tasks $\mathcal{T}_{i}^{\text{train}}$, the test dataset $D^{\text{test}}$ is divided into a collection of meta-testing tasks $\mathcal{T}_{i}^{\text{test}}$. Each task $\mathcal{T}_{i}$ is associated with two disjoint sets, a support set $S_{i}$ and a query set $Q_{i}$ \cite{Rajeswaran2019}. In meta-training, the objective is to learn a model $f_{\theta}$, parameterized by $\theta$, that can generalize across tasks \cite{Rajeswaran2019}. The performance on a task is specified by a loss function $\mathcal{L}$ which measures the error between correct labels and those predicted by $f_{\theta}$. Then, $\mathcal{L}_{D}(\theta)$ denotes the loss on dataset $D$ based on $f_{\theta}$, defined as a function of the parameters $\theta$,
\begin{align}
    \mathcal{L}_{D}(\theta) &:= \frac{1}{|D|}\sum_{(x,y) \in D} \mathcal{L}(f_{\theta}(x), y).
\end{align}
The support set $S^{\text{train}}_{i}$ is used for adapting the model on a task $\mathcal{T}^{\text{train}}_{i}$, while the query set $Q^{\text{train}}_{i}$ is used to evaluate this adaptation \cite{Rajeswaran2019}. The generalization performance on a meta-testing task $\mathcal{T}^{\text{test}}_{i}$ is assessed by the loss on the query set $Q^{\text{test}}_{i}$ after adaptation on the support set $S^{\text{test}}_{i}$ \cite{Rajeswaran2019}. The meta-learning objective can be formulated as 
\begin{align} \label{general meta-learning problem}
    \theta^{\star} &:= \underset{\theta} {\arg\max} \ \underset{\substack{\mathcal{T}\sim p(\mathcal{T}) \\ \mathcal{T}=(S, Q) }}{\mathbb{E}} \sum_{(x,y) \in Q} \log p_{\theta}(y|x, S),
\end{align}
where $p(\mathcal{T})$ denotes a distribution over tasks. A standard assumption in meta-learning is that both meta-training and meta-testing tasks are drawn from the same task distribution \cite{Finn2017}. 

The optimization-based meta-learning approach frames the meta-training process as a bi-level optimization problem. The goal is to learn meta-parameters that produce good task-specific parameters after an adaptation procedure $\mathcal{A}lg$ \cite{Rajeswaran2019}. $\mathcal{A}lg$ corresponds to an algorithm that computes task-specific parameters $\phi_{i}$ using a set of meta-parameters $\theta$ and the support set $S_{i}$ of a task $\mathcal{T}_{i}$ \cite{Rajeswaran2019}. Given a meta-batch $(\mathcal{T}_{i})_{i=1}^{N}$ of meta-training tasks $\mathcal{T}_{i}=(S_{i}, Q_{i})$, the meta-learning problem in Eq. \ref{general meta-learning problem} can be formalized as 
\begin{align}
    \theta^{\star} &:= \underset{\theta} {\arg\max} \ \frac{1}{N} \sum_{i=1}^{N} \mathcal{L}_{Q_{i}}(\phi^{\star}_{i}(\theta)) \label{bi-level meta-learning: outer level}\\ \text{s.t.} \ \phi^{\star}_{i}(\theta) &:= \mathcal{A}lg (\theta, S_{i}).\label{bi-level meta-learning: inner level}
\end{align}
Since $\mathcal{A}lg (\theta, S_{i})$ is typically interpreted as solving explicitly \cite{Finn2017} or implicitly \cite{Ravi2017} an underlying optimization problem, this is viewed as a bi-level optimization problem \cite{Rajeswaran2019}.


\subsection{One-Class Domain Adaptation via Meta-Learning}
In the next section, we will introduce the One-Class Domain Adaptation (OC-DA) setting. In OC-DA, data is distributed across multiple domains. The goal of OC-DA is to learn a model that is able to quickly adapt to a new domain using a few examples from only \textit{one} class. We will refer to this class as the normal class $n$. The model is a multi-class classifier, capable of differentiating between normal class examples and examples of multiple other classes using only a few normal class examples. OC-DA allows addressing distribution shift challenges in real-world applications where only normal operational data is available for model adaptation in new environments.

\begin{figure}[!htbp]
	\centerline{\includegraphics[width=0.5\textwidth]{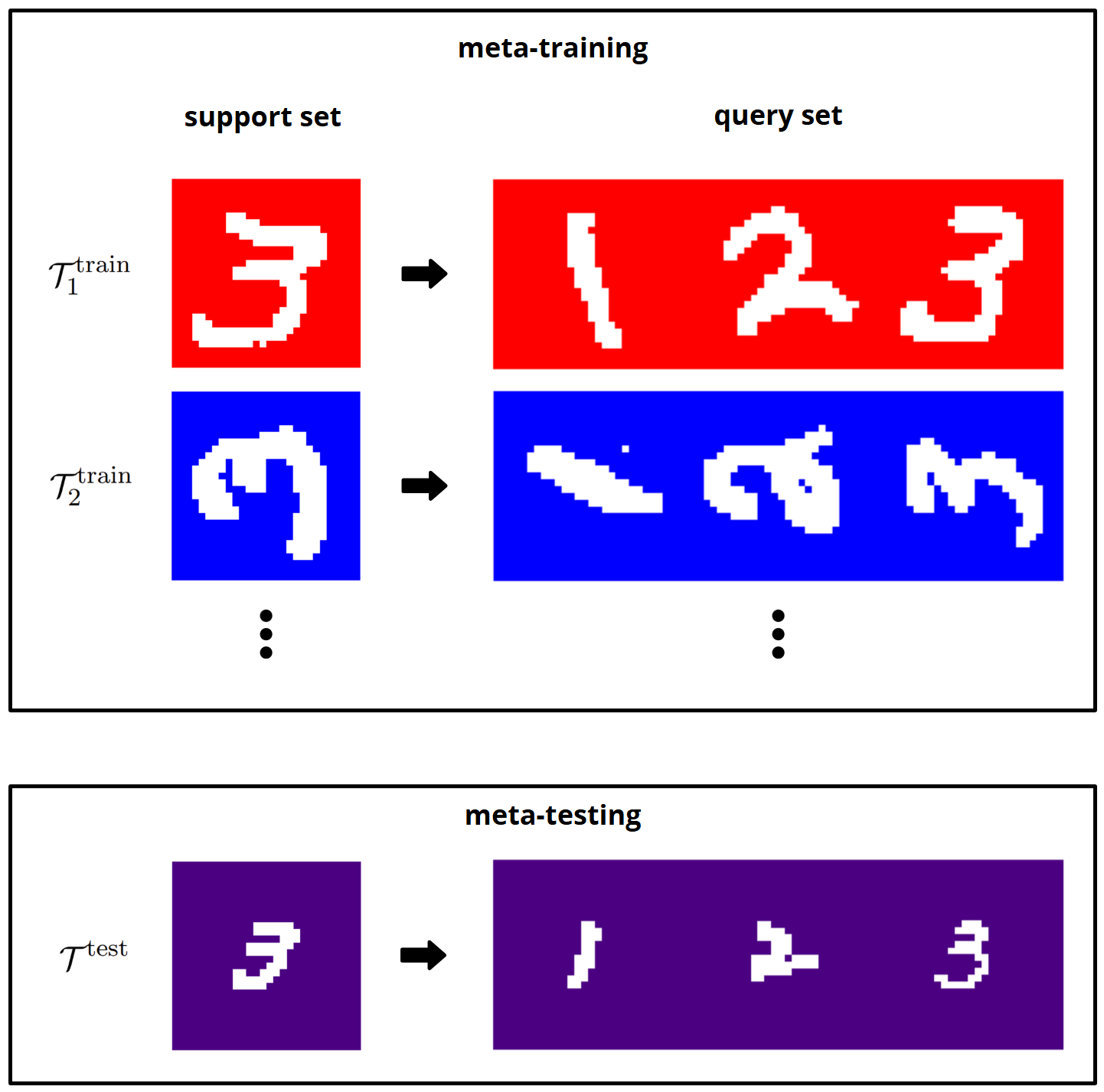}}
	\caption{Example of a 3-way 1-shot learning task in the OC-DA setting.}
	\label{OurTaskSetup}
\end{figure}

We take a meta-learning approach to tackle the challenge of OC-DA. While recent meta-learning algorithms are designed to effectively address generalization across \textit{diverse} learning tasks \cite{Koch2015,Snell2017,Ravi2017,Finn2017}, that is, learning tasks with diverse label spaces, OC-DA focuses on generalization from normal class examples to examples of multiple other classes of a \textit{single} learning task across diverse domains. The bi-level optimization framework of optimization-based meta-learning enables direct optimization for one-class adaptation across domains. Meta-training tasks are sampled from the source domains, while meta-testing tasks are sampled from the target domains. Meta-learning algorithms typically sample tasks where both the support and query sets are class-balanced. We propose a task sampling strategy that aligns the task setup during meta-training with that of meta-testing. Thus, the support set is restricted to normal class examples, while the query set remains class-balanced in meta-training tasks. This strategy requires the domain-specific parameters obtained by performing the adaptation procedure $\mathcal{A}lg$ using a few normal class examples to improve performance on a class-balanced dataset within the domain. Fig. \ref{OurTaskSetup} illustrates $3$-way $1$-shot learning in the OC-DA setting on the Rainbow-MNIST dataset. 

. 

\subsection{One-Class Domain Adaptation via Model-Agnostic Meta-Learning (OC-DA MAML)} 
We apply our proposed task sampling strategy to the MAML algorithm \cite{Finn2017}. The key idea of model-agnostic meta-learning (MAML) \cite{Finn2017} is to learn an initialization for the parameters $\theta$ of a neural network $f_{\theta}$. MAML ensures that with only a few gradient descent steps on a few examples of a new task, the model can achieve good task-specific parameters \cite{Finn2017}. 

In the bi-level optimization view, the inner-level and outer-level problems are both minimized via stochastic gradient descent. The inner-level algorithm $\mathcal{A}lg (\theta, S_{i})$ in Eq. \ref{bi-level meta-learning: inner level} corresponds to a few gradient descent steps on the support set $S_{i}$ initialized at meta-parameters $\theta$. For simpler notation, only one gradient descent step is used in the following. By this, MAML explicitly optimizes for parameters $\theta$ such that one gradient descent step on a small number of examples from a new task $\mathcal{T}_{i}$ produces good task-specific parameters $\phi_{i}(\theta)$, 
\begin{align}
    \mathcal{A}lg (\theta, S_{i}):= \phi_{i}(\theta) := \theta - \alpha \ \nabla \mathcal{L}_{S_{i}}(\theta). \label{MAML: inner-level}
\end{align}
The outer-level problem in Eq. \ref{bi-level meta-learning: outer level} is solved by performing a gradient descent step on the corresponding query sets,  
\begin{align}
    \theta^{\star}:=\theta - \beta \ \nabla \Big(\frac{1}{N} \sum_{i=1}^{N}  \mathcal{L}_{Q_{i}}\big(\phi_{i}(\cdot)\big)\Big)(\theta) =\theta - \frac{\beta}{N} \sum_{i=1}^{N}  \nabla \big(\mathcal{L}_{Q_{i}}\circ \phi_{i}\big)(\theta). \label{MAML: outer-level}
\end{align}
Note that the outer-level loss in Eq. \ref{bi-level meta-learning: outer level} is computed based on the task-specific parameters $\phi$, however, the outer-level optimization is performed over the meta-parameters $\theta$ \cite{Finn2017}. Therefore, MAML can be viewed as a form of transfer learning with many tasks, where parameters are differentiated through the fine-tuning process, i.e. the adaptation procedure $\mathcal{A}lg$ \cite{Nichol2018}. 

The OC-DA MAML algorithm for the meta-training process is described in Algorithm \ref{alg: Ours}. The neural network $f_{\theta}$ is randomly initialized. In each iteration step, a random batch of domains $I \subset \mathcal{E}$ is selected. For each domain $i \in I$, a $N$-way $K$-shot learning task is sampled according to our task sampling strategy: the support set $S_{i}$ contains $K$ normal class examples, while the query set $Q_{i}$ is a class-balanced dataset with $K$ examples per class. Then, the domain-specific parameters $\phi_{i}(\theta)$ are computed by performing a gradient descent step on the support set. Finally, the meta-parameters $\theta$ are updated via a gradient descent step on the query set. Note that this includes differentiating through the fine-tuning process on the support set. By this, we explicitly optimize the meta-parameters $\theta$ such that performing $k$ gradient descent steps on a few normal class examples yields effective domain-specific parameters $\phi_{i}(\theta)$.

\begin{algorithm}\small
	\DontPrintSemicolon
	\caption{MAML for One-Class Domain Adaptation (OC-DA MAML)}
	\SetAlgoLined
	\label{alg: Ours}
	\textbf{Required:} 
	model $f_{\theta}$ parameterized by $\theta$, loss function $\mathcal{L}$, inner learning rate $\alpha$, outer learning rate $\beta$, set of class labels $C$, normal class $n \in C$, number of inner gradient descent steps $k$
	
	Randomly initialize $ \theta_{0}$ \\
	
    \While{not done}{
    Sample batch of domains $I \subset \mathcal{E}$
    
        \For{$i \in I$}{
            
            \text{Sample $K$ examples ${S}_{i}:=\{(x_{j}, n)\}_{j=1}^{K}$ from $D_{i}$} \\

            \For{$c \in C$}{
                \text{Sample $K$ examples $Q^{c}_{i}:=\{(x_{j}, c)\}_{j=1}^{K}$ from $D_{i}$}
                }
            $Q_{i}:= \underset{c \in C}{\cup}Q^{c}_{i}$ \\
            \text{Compute domain-specific parameters via gradient descent step:} \\
            $\phi_{i}(\theta_{t}) := \theta_{t} - \alpha \nabla \mathcal{L}_{S_{i}}(\theta_{t})$
            
        }
        \text{Perform meta-update via gradient descent:}\\
        $\theta_{t+1} := \theta_{t} - \frac{\beta}{|I|} \sum_{i \in I}  \nabla_{\theta_{t}} \mathcal{L}_{Q_{i}}\big(\phi_{i}(\theta_{t})\big)$
    }
\end{algorithm}

\subsection{Analysis of OC-DA MAML}
In the following section, we present an analysis to provide a deeper understanding of our proposed task sampling strategy and its underlying motivation. Specifically, we will use a Taylor series expansion to approximate the gradient of the MAML algorithm, $g(\theta):= \nabla_{\theta} \mathcal{L}_{Q} \big(\phi(\theta)\big)$ \cite{Nichol2018}. This analysis will focus on two key components: the first term minimizes the loss on the query set, while the second term maximizes the inner product between the gradients of the support and query loss \cite{Nichol2018}. 

Using the second-order Taylor series expansion of $\nabla \mathcal{L}_{Q}$ around $\theta$, for small $\alpha>0$, the MAML gradient can be approximated as \cite{Nichol2018}

\begin{align*}
    g(\theta) &= \nabla (\mathcal{L}_{Q} \circ \phi)(\theta) \\&= \big(I - \alpha \nabla^{2} \mathcal{L}_{S}(\theta)\big) \hspace{1mm} \nabla \mathcal{L}_{Q} \big(\phi(\theta)\big) \\&= \big(I - \alpha \nabla^{2} \mathcal{L}_{S}(\theta)\big) \hspace{1mm} \big(\nabla \mathcal{L}_{Q}(\theta) -\alpha \nabla^{2} \mathcal{L}_{Q}(\theta) \nabla \mathcal{L}_{S}(\theta) + \mathcal{O}(\alpha^{2}) \big)\\&= \nabla \mathcal{L}_{Q}(\theta) - \alpha \nabla^{2} \mathcal{L}_{Q}(\theta) \nabla \mathcal{L}_{S}(\theta) - \alpha \nabla^{2} \mathcal{L}_{S}(\theta) \nabla \mathcal{L}_{Q}(\theta) + \mathcal{O}(\alpha^{2})\\&=  \nabla \big( \mathcal{L}_{Q} - \alpha \nabla \mathcal{L}_{Q} \cdot \nabla \mathcal{L}_{S}\big)(\theta) + \mathcal{O}(\alpha^{2}).
\end{align*}

Then, the meta-update can be written as
\begin{align*}
    \theta_{t+1} &= \theta_{t} - \frac{\beta}{|I|} \sum_{i \in I}  \nabla_{\theta_{t}} \mathcal{L}_{Q_{i}} \big(\phi_{i}(\theta_{t})\big) \\& = \theta_{t} - \frac{\beta}{|I|} \sum_{i \in I} \nabla_{\theta_{t}} \big(\mathcal{L}_{Q_{i}}(\theta_{t}) - \alpha \nabla \mathcal{L}_{Q_{i}}(\theta_{t}) \cdot \nabla \mathcal{L}_{S_{i}}(\theta_{t})\big) + \mathcal{O}(\alpha^{2}).
\end{align*}

While the first term minimizes the loss on the query set, the second term maximizes the inner product between the gradients of the support and query loss \cite{Nichol2018}. When this inner product is positive, performing a gradient descent step on one dataset improves performance on the other \cite{Nichol2018}. The MAML algorithm maximizes this inner product, thereby finding update directions that minimize both the support and query loss \cite{Li2017}. Consequently, a gradient descent step on the support set improves performance on the query set, and thus, enables within-task generalization \cite{Nichol2018}. In the context of OCDA-MAML, this implies that gradient descent steps on normal class examples not only improve performance on normal data but also improve performance on data including other classes. At meta-testing time, when the model only has access to normal class examples, fine-tuning on a few normal class examples will produce good domain-specific parameters while avoiding over-fitting.

\section{Experiments}
\subsection{Experimental Setup}
In the next chapter, we present our experiments on distribution shifts and one-class domain adaptation. We implemented both standard learning and meta-learning methods. In the context of standard learning, we illustrate the performance gap between source and target domains. For meta-learning, we implemented the MAML algorithm and our OC-DA MAML algorithm in the OC-DA setting. We aim to show that the MAML algorithm is not tailored to OC-DA settings and demonstrate the effectiveness of our task sampling strategy in OC-DA MAML, enabling rapid one-class adaptation. To ensure the reliability of our results, all results were averaged over three runs with different seeds. In the implementation of the MAML and OC-DA MAML algorithm, we used the 'learn2learn' library \cite{Arnold2020}, a library for meta-learning research that provides low-level routines built on top of PyTorch for few-shot learning and differentiable optimization (e.g. automatic differentiation through the meta-updates in MAML).

In Rainbow-MNIST, we adopt the approach of Yao et al. \cite{Yao2022} by employing a convolutional neural network that consists of four convolutional blocks, followed by a linear layer and softmax activation function. Each convolutional block is designed with a two-dimensional convolutional layer with $32$ filters of size $3 \times 3$, a batch normalization layer, a ReLU activation function, and a two-dimensional max-pooling layer of size $2 \times 2$.  In the Centrifugal-Pumps dataset, we use the same model architecture as Frikha et al. \cite{Frikha2021}, that is, a convolutional neural network with three convolutional blocks, a linear layer and a softmax activation function. Each block consists of a one-dimensional convolutional layer with $32$ filters of size $5$, a one-dimensional max-pooling layer of size $2$, and a ReLU activation function. 

We empirically evaluate our approach on two datasets. First, we utilize a well-established meta-learning benchmark, the Rainbow-MNIST dataset \cite{Finn2019} and adapt it to our OC-DA setting. Second, we assess our approach using a real-world dataset of vibration-based sensor readings recorded by four centrifugal pumps across multiple domains (Centrifugal-Pumps dataset), demonstrating its robustness and applicability in real-world applications. The goal of our experimental evaluation is to answer the following questions: (1) Can we leverage domain-specific information present in one class for efficient domain adaptation? (2) Can OC-DA MAML enable rapid one-class adaptation in the context of domain adaptation tasks? (3) How does OC-DA MAML compare to the original MAML algorithm in the OC-DA setting?

\subsection{Results}
\begin{figure}[!htbp]
	\centerline{\includegraphics[width=1\textwidth]{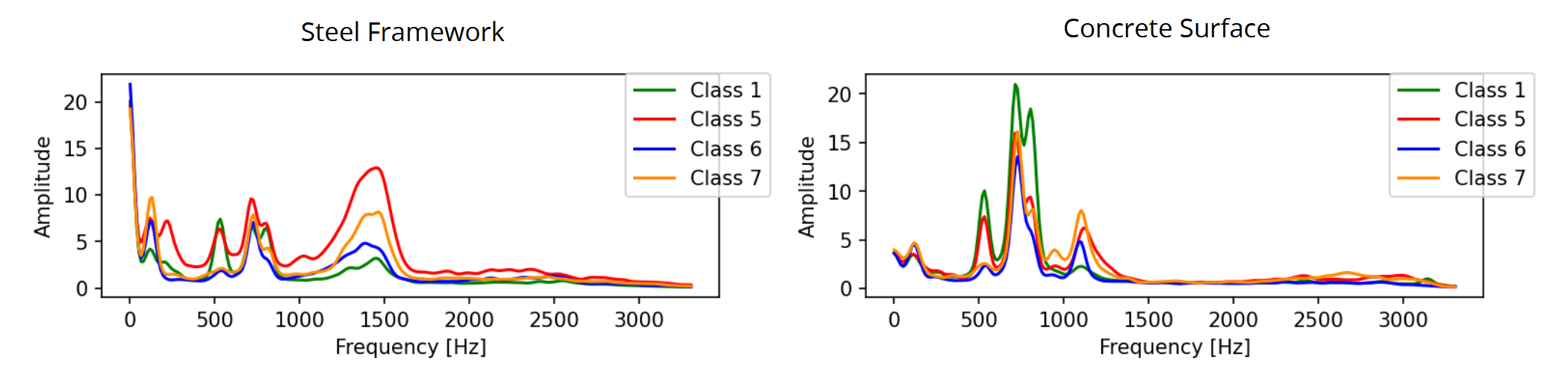}}
	\caption{Visualization of distribution shifts in the Centrifugal-Pumps dataset, showing the average amplitude [mm] of FFT vibration signals per frequency [Hz] for each class (1-normal, 5-cavitation, 6-hydraulic blockage, 7-dry running), recorded by the same pump operated within a steel framework vs. on a concrete surface.}
	\label{AveragePerClass}
\end{figure}

Fig. \ref{AveragePerClass} illustrates differences in the data distribution across two different domains in the Centrifugal-Pumps dataset. The plots show the average amplitude of FFT vibration signals for each class (1-normal, 5-cavitation, 6-hydraulic blockage, 7-dry running). The data was recorded by the same pump placed on two different surfaces: within a steel framework and on a concrete surface. We observe significant differences in the data distribution. Empirical experiments will later confirm that these visual differences in the data distribution present challenges for classification models and lead to a substantial performance drop, see Table \ref{Table: Centrifugal-Pumps}. 

\begin{figure}[!htbp]
	\centerline{\includegraphics[width=1\textwidth]{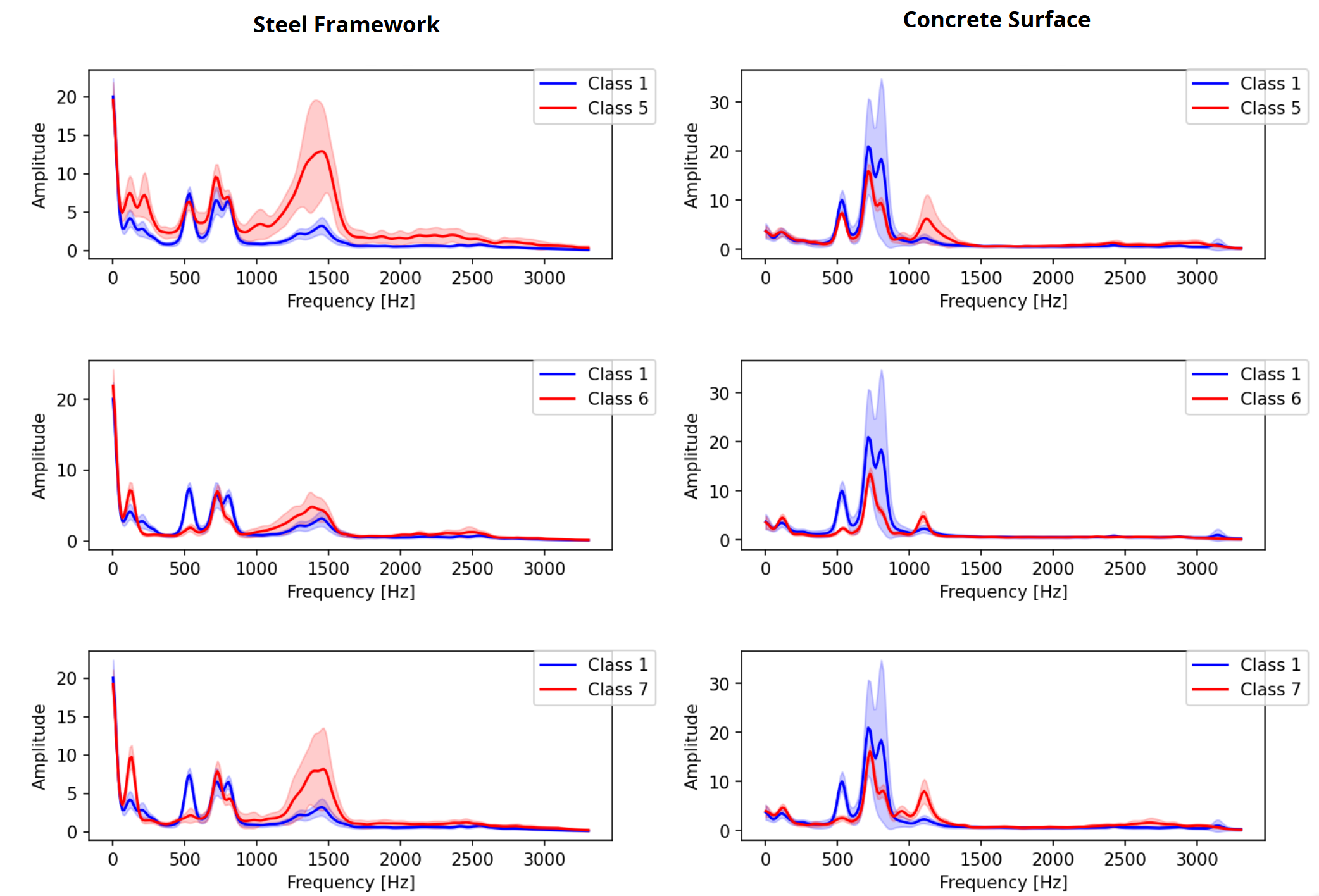}}
	\caption{Visualization of domain-specific information in the Centrifugal-Pumps dataset, showing the average amplitude [mm] of the FFT vibration signals per frequency [Hz] with a $95\%$ confidence interval. Each anomalous class (5-cavitation, 6-hydraulic blockage, 7-dry running) is compared to the normal class (1-normal class), recorded by same pump operated within a steel framework vs. on a concrete surface.}
	\label{DataDistribution}
\end{figure}

Fig. \ref{DataDistribution} illustrates domain-specific patterns in the Centrifugal-Pumps dataset. The plots show the $95\%$ confidence interval of the FFT vibration signals in the frequency spectrum. We observe that the normal class encodes domain-specific information, in the sense that the vibration signals of the anomalous classes follow the pattern of the normal class. This observation motivates the question whether the information present in normal data can be leveraged for adaptation to another domain.

In the context of distribution shifts, accurately measuring the performance drop caused by these shifts is challenging. The machine learning literature distinguishes between in-distribution (ID) and out-of-distribution (OOD) performance \cite{Koh2021}. ID performance is measured by training a model on the training distribution and evaluating its performance on held-out data from the training distribution \cite{Koh2021}. This metric provides insight into how well the model generalizes on the training distribution. In contrast, OOD performance is measured by the model's performance on data from the test distribution \cite{Koh2021}. This metric provides insight into how well the model generalizes to data outside its training distribution. 

Table \ref{Table: Distribution Shift} presents the experimental results on the Rainbow-MNIST dataset, demonstrating the performance gap between source and target domains. We observe high ID performance and low ODD performance. Additionally, the ID-test performance significantly outperforms the ODD performance, indicating that the target domains are not intrinsically more challenging and the performance gap between source and target domains is indeed due to distribution shifts. 

\begin{table}[!htbp]\small
    \begin{center}
        \begin{tabular}{|lrrr|}
            \hline
              Metric & Training dataset & Test dataset & Rainbow-MNIST \\
             \hline
            ID & source  & source  & 83.93 \\
            OOD & source  & target  & 13.5 \\
            ID-test & target  & target  & 72.77 \\
            \hline
        \end{tabular}
        \caption{Experimental results of standard learning on source and target domains on Rainbow-MNIST. Note that the ID metric is evaluated on held-out data from each source domain, the ID-test metric is evaluated on held-out data from each target domain.}
        \label{Table: Distribution Shift}
    \end{center}
\end{table}

Table \ref{Table: Rainbow-MNIST} presents the experimental results of the meta-learning approaches on the Rainbow-MNIST dataset in the OC-DA setting, showing the average accuracy [$\%$] computed on the target domain datasets. To adapt the Rainbow-MNIST dataset to the OC-DA setting, we designate one class as the normal class. We conducted experiments for each class, using two different values for $K$. The hyperparameters are presented in more detail in Table \ref{Table: Hyperparameters}. We observe that performance improves with larger values of $K$. Furthermore, the results demonstrate the effectiveness of our task sampling strategy in OC-DA MAML, significantly outperforming MAML with classical $K$-shot learning tasks. Note that OC-DA MAML achieves results on the target domains that are comparable to the ID performance on the source domains, as shown in Table \ref{Table: Distribution Shift}.

\begin{table}[!htbp]\small
    \begin{center}
        \begin{tabular}{|l|cc|cc|}
            \hline
            & \multicolumn{2}{c|}{$K=1$} & \multicolumn{2}{c|}{$K=3$}\\
            Class $n$ & MAML & OC-DA MAML & MAML & OC-DA MAML \\
            \hline
            0 & 41.33 & \textbf{82.17} & 44.2 & \textbf{86.07} \\
            1 & 36.9 & \textbf{80.87} & 41.63 & \textbf{81.77} \\
            2 & 32.33 & \textbf{80.77} & 39.53 & \textbf{89.57} \\
            3 & 20.03 & \textbf{84.93} & 31.1 & \textbf{92.67} \\
            4 & 37.07 & \textbf{84.63} & 43.87 & \textbf{91.1} \\
            5 & 32.37 & \textbf{82.3} & 31.3 & \textbf{87.87}\\
            6 & 37.23 & \textbf{86.1} & 31.4 & \textbf{92.47} \\
            7 & 35.87 & \textbf{86.23} & 39.23 & \textbf{93.07}\\
            8 & 41.73 & \textbf{82.73} & 40.1 & \textbf{86.73}\\
            9 & 41.4 & \textbf{78.87} & 41.43 & \textbf{82.33}\\
            \hline
            Average & 35.6 & \textbf{82.9} & 38.4 & \textbf{88.4}\\
            \hline
        \end{tabular}
        \caption{Experimental results of meta-learning approaches on Rainbow-MNIST in OC-DA setting: average accuracy [\%] on target domains after adaptation on $K$ normal class examples.}
        \label{Table: Rainbow-MNIST}
    \end{center}
\end{table}

Table \ref{Table: Centrifugal-Pumps} shows the experimental results on the Centrifugal-Pumps dataset. In each source and target domain combination, note that there are $16$ source domains and $4$ target domains, as outlined in Section \ref{Data}. For the distribution shift metrics, we down-sampled each domain dataset by reducing the number of examples per class to match the size of the minority class. The table shows the average accuracy [$\%$] computed on the respective test dataset. We observe a performance drop from the target to the source domains. The OC-DA MAML algorithm approximates the ID-metric and outperforms the standard MAML algorithm. 

\begin{table}[!htbp]\small
    \begin{center}
        \begin{tabular}{|ll|cc|cc|}
            \hline
            & & & & \multicolumn{2}{c|}{K=2}\\
            Source domains & Target domains & ID & OOD & MAML & OCDA-MAML \\
            \hline
            $P_{1}, P_{2}, P_{3}$ 'steel' & $P_{4}$ 'concrete' & 96.2 & 80.3 & 87.73 & \textbf{89.63}\\
            $P_{1}, P_{2}, P_{4}$ 'steel' & $P_{3}$ 'concrete' & 94.77 & 89.56 & 74.03 & \textbf{93.7}\\
            $P_{1}, P_{3}, P_{4}$ steel & $P_{2}$ 'concrete' & 94.86 & 79.73 & 76.17& \textbf{91.2}\\
            $P_{2}, P_{3}, P_{4}$ 'steel' & $P_{1}$ 'concrete' & 94.23 & 77.57 & 91.53 & \textbf{92.43}\\
            $P_{1}, P_{2}, P_{3}$ 'concrete' & $P_{4}$ 'steel' & 92.63 & 71.4 & 92.47& \textbf{93.5}\\
            $P_{1}, P_{2}, P_{4}$ 'concrete' & $P_{3}$ 'steel' & 93.93 & 88.9 & 86.9& \textbf{93.1}\\
            $P_{1}, P_{3}, P_{4}$ 'concrete' & $P_{2}$ 'steel' & 95.07 & 89.53 & 81.4& \textbf{98.6}\\
            $P_{2}, P_{3}, P_{4}$ 'concrete' & $P_{1}$ 'steel' & 94.67 & 80.83 & 83.8 & \textbf{90.47}\\
            \hline
            \multicolumn{2}{|l|}{Average} & 95.5 & 82.2 & 84.3 & \textbf{92.8} \\
            \hline
        \end{tabular}
        \caption{Experimental results on the Centrifugal-Pumps dataset. ID: accuracy [\%] on held-out dataset from source domains; OOD: average accuracy [\%] on target domains; MAML/OC-DA MAML: average accuracy [\%] on target domains after adaptation on $K$ normal class examples 
}
        \label{Table: Centrifugal-Pumps}
    \end{center}
\end{table}

Table \ref{Table: Hyperparameters} shows the hyerparameters we used in meta-learning. We used the Adam optimizer with corresponding learning rates $\alpha$, $\beta$ and a weight decay of $1e-5$. In all experiments, we used the categorical cross entropy loss.
\begin{table}[!htbp]\small
    \begin{center}
        \begin{tabular}{|lrr|}
            \hline
            Hyperparameter & Rainbow-MNIST & Centrifugal-Pumps \\
            \hline
            Input size & $28 \times 28$ & $1 \times 256$\\
            Meta-batch size $|I|$  & 4 & 2 \\
            Meta-training iterations & 30,000 & 20,000\\
            Inner gradient descent steps $k$ & 1 & 1 \\
            Inner learning rate $\alpha$ & 0.01 & 0.01 \\
            Outer learning rate $\beta$ & 0.001 & 0.001 \\
            $N$ (Classes per task) & 10 & 5 \\
            $K$ (Shots per class) & $\{1, 3\}$ & 2 \\
            \hline
        \end{tabular}
        \caption{Hyperparameters in MAML and OC-DA MAML.}
        \label{Table: Hyperparameters}
    \end{center}
\end{table}

\section{Conclusion and Future Work}
We proposed a task sampling strategy to adapt any bi-level meta-learning algorithm to the OC-DA setting, and introduced the OC-DA MAML algorithm. We provided a theoretical analysis of the OC-DA MAML meta-update, demonstrating that OC-DA MAML explicitly optimizes for meta-parameters that enable generalization from one class to the other classes within a domain, and thus, one-class adaptation across domains. The empirical results support these theoretical observations. We evaluated the OC-DA MAML algorithm on a meta-learning benchmark and demonstrated its robustness in real-world applications using a dataset of vibration-based sensor readings recorded by centrifugal pumps in diverse environments. The OC-DA MAML algorithm consistently outperforms the standard MAML algorithm for all source and target domain combinations. We conclude that it is possible to leverage domain-specific information present in one class for efficient domain adaptation. The proposed task sampling strategy in bi-level meta-learning enables generalization from one class to other classes within a domain, and thus, one-class adaptation across domains.

Despite the growing demand for machine learning in industry, transferring models from laboratory settings to real-world deployments is an open challenge. While we achieved promising results on a real-world dataset, the centrifugal pumps were operated within a laboratory setting. It would be interesting to explore the performance of OC-DA MAML in actual deployments, including a large number of pumps in diverse environments. Additionally, we only considered centrifugal pumps of the same type and size. In reality, it is common to encounter pumps assembled according to a modular system. Future research could therefore explore generalizing the models not only to different pumps in new environments but also to entirely new pump configurations.

%
%
%
%

\appendix 
\section{Appendix}
\subsection{Data}\label{Data}
Rainbow-MNIST \cite{Finn2019} is a variant of the MNIST dataset that provides 56 domains, where each domain corresponds to a combination of a background color ('red', 'orange', 'yellow', 'green', 'blue', 'indigo', 'violet'), degree of rotation ($0^{\circ}, 90^{\circ}, 180^{\circ}, 270^{\circ}$) and scale size ('full', 'half'). Following the approach of Finn et al. \cite{Finn2019}, we split the MNIST dataset into $56$ class-balanced sub-datasets, each containing $1000$ examples, and applied the corresponding domain transformation to each sub-dataset. Support and query tasks are sampled randomly from these domain datasets. To adapt the meta-learning benchmark to the OC-DA setting, we designate one class in the original dataset as the normal class. We split the domains into 40/8/8 domains for meta-training/validation/testing.

The Centrifugal-Pumps dataset comprises sensor readings recorded from four identical centrifugal pumps in a controlled laboratory environment, each equipped with IoT sensors recording vibration data \cite{Holly2021}. In order to create a diverse dataset with multiple domains, we simulated varying environmental influences by placing the pumps on different surfaces. For each measurement round, a pump was placed either on a concrete surface or within a steel framework and operated under multiple conditions, including normal operational data, idle state data and three anomalous conditions: hydraulic blockage, dry running and cavitation. The dataset provides $32$ domains. We split the domains into multiple combinations of source and target domains, each combination corresponding to a transfer to a new pump within an unknown environment. Specifically, we split the domains into $12/4/4$ domains for meta-training/validation/testing. This setup results in a total of eight source and target domain combinations. 

\subsection{Task Setup}
In this section, we detail our experimental setup. We divide the source domains into training and validation domains. During meta-training, meta-training tasks are sampled from the training domains, meta-validation tasks are sampled from the validation domains. Meta-training is stopped either by reaching the maximal number of iterations or through early stopping. In MAML, the meta-training and validation tasks are $K$-shot learning tasks, including $K$ examples per class for both the support and query sets. In the OC-DA MAML sampling strategy, only $K$ examples of the normal class are sampled for the support set, see Table \ref{Table: Task Setup}. In each target domain $i \in \mathcal{E}^{\text{target}}$, the model is adapted on $K$ normal examples (meta-testing support set) and evaluated on a class-balanced dataset (meta-testing query set). In Rainbow-MNIST, the domains are class-balanced by design. In Centrifugal-Pumps, we down-sampled the target domain datasets $D_{i}$ by reducing the number of examples per class to match the size of the minority class. Here, $N^{i}_{c}$ denotes the number of examples in $D_{i}$ that belong to class $c$, $N^{i}_{c}:=|\{j| (x_{j}, y_{j}) \in D_{i}, y_{j}=c\}|$. Note that the MAML algorithm is designed for classical meta-learning settings, where the task setup typically differs from the task setup in the OC-DA setting, as outlined in Table \ref{Table: Task Setup}. In classical meta-learning settings, the support and query sets of meta-training, validation, and testing tasks are $K$-shot learning tasks. In contrast, in the OC-DA setting, the support set of meta-testing tasks is limited to $K$ shots of the normal class, as detailed in Table \ref{Table: Task Setup}. Furthermore, evaluating the model's performance using only $K$-shots per class in the target domain is impractical for domain adaptation. Therefore, we assess the model's performance on a down-sampled dataset, rather than a dataset with $K$-shots per class.

\begin{table}[!htbp]\small
    \begin{center}
        \begin{tabular}{|ll|cc|c|}
            \hline
             & & \multicolumn{2}{c|}{Shots per class $c$ (support)} & Shots per class $c$ (query) \\
             & & $c=n$ & $c \in C\setminus \{n\}$ & $c \in C$\\
             \hline
             \multirow{3}{*}{MAML}& Meta-training & $K$ & $K$ & $K$ \\
             & Meta-validation & $K$ & $K$ & $K$ \\
             & Meta-testing & $K$ & $0$ & $\underset{c \in C}{\min} N^{i}_{c}$ \\
            \hline
            \multirow{3}{*}{OC-DA MAML}& Meta-training & $K$ & $0$ & $K$ \\
             & Meta-validation & $K$ & $0$ & $K$ \\
             & Meta-testing & $K$ & $0$ & $\underset{c \in C}{\min} N^{i}_{c}$ \\
            \hline
        \end{tabular}
        \caption{Task sampling strategy of MAML vs. OC-DA MAML.}
        \label{Table: Task Setup}
    \end{center}
\end{table}

\subsection{Data Splits}
In standard learning, we split the data into training and validation data and applied early stopping. For the ID metric, we collected, shuffled and split the data in the source domains into a training, validation and test dataset. The ID metric is computed as the accuracy [$\%$] on this test dataset. For the ODD metric, we collected, shuffled and split the data in the source domains into a training and validation dataset. The ODD metric is computed as the average accuracy [$\%$] over the class-balanced target domain datasets. For the ID-test metric, we collected, shuffled and split the data in the target domains into a training, validation and test dataset. The ID-test metric is computed as the accuracy [$\%$] on this test dataset.

\end{document}